\documentclass[10pt,twocolumn,letterpaper]{article}

\usepackage{cvpr}
\usepackage{times}
\usepackage{epsfig}
\usepackage{graphicx}
\usepackage{amsmath}
\usepackage{amssymb}
\usepackage{subfig}
\usepackage{algorithm}
\usepackage{algorithmic}
\usepackage{multirow}
\usepackage{caption}
\usepackage{amsthm}
\usepackage{authblk}
\usepackage[breaklinks=true,bookmarks=false]{hyperref}

\cvprfinalcopy 

\def\cvprPaperID{****} 

\begin{document}

\title{Moving Object Detection in Video Using Saliency Map and Subspace Learning}


\author[a]{Yanwei Pang \thanks{pyw@tju.edu.cn}}
\author[a]{Li Ye}
\author[b]{Xuelong Li}
\author[a,c]{Jing Pan}

\affil[a]{School of Electronic Information Engineering, Tianjin University, Tianjin 300072, China}
\affil[b]{Institute of Optics and Precision Mechanics, Chinese Academy of Sciences, Xi'an 710119, China}
\affil[c]{School of Electronic Engineering, Tianjin University, Tianjin 300222, China}

\maketitle
\begin{abstract}
Moving object detection is a key to intelligent video analysis. On the one hand, what moves is not only interesting objects but also noise and cluttered background. On the other hand, moving objects without rich texture are prone not to be detected. So there are undesirable false alarms and missed alarms in many algorithms of moving object detection. To reduce the false alarms and missed alarms, in this paper, we propose to incorporate a saliency map into an incremental subspace analysis framework where the saliency map makes estimated background has less chance than foreground (i.e., moving objects) to contain salient objects. The proposed objective function systematically takes account into the properties of sparsity, low-rank, connectivity, and saliency. An alternative minimization algorithm is proposed to seek the optimal solutions. Experimental results on the Perception Test Images Sequences demonstrate that the proposed method is effective in reducing false alarms and missed alarms.
\end{abstract}

\section{Introduction}

Object detection is the basis of intelligent video analysis. Generally, object recognition, action and behavior recognition, and tracking rely on the detected objects. In a sequence of images, there are both moving and static objects. In this paper, the focus is on detecting moving objects in a video. 
Moving object detection is related to but also different from class-specific object detection and general salient object detection. Pedestrian detection, face detection, and hand detection are instances of class-specific object detection. The task of moving object detection is to detect semantically meaningful moving objects. Predefined classes of moving objects should be detected by a moving object detection algorithm. Moreover, other semantically meaningful objects should also be detected even though their classes are not pre-defined. Examples of meaningless moving objects include water ripples, waving trees (leafs), shadows, noisy data, and the one caused by variations of illumination. However, the moving object detection algorithm relying merely on motion information is prone to incorrectly classify such meaningless moving objects as meaningful ones. The corresponding error is called false alarms. But a salient object detection algorithm tends to correctly discard the meaningless objects. Hence, in this paper, we propose to incorporate the output (i.e., saliency map) of a salient object algorithm into a subspace analysis based objective function so that the problem of false alarms can be alleviated. 
It is noted that our method is also capable of alleviating the problem of missed alarms. Existing moving object detection algorithms tend to classify flat regions (i.e., textureless regions) inside an object and moving regions with similar appearance (texture) to background as static background and thus such regions may be missed. State-of-the-art salient object detection algorithm can output large value of saliency map at such regions. Utilizing the saliency map, our method has ability to classify such regions as foreground. 
In summary, we present an objective function that unifies subspace analysis of background and saliency map. The objective function consists of four terms: saliency map, sparsity, connectivity, and low-rank. An alternative minimization algorithm is proposed to find the optimal solution. The significant advantage compared to previous subspace based approaches is that saliency map is used to guide the result to have less false and missed alarms. The proposed method is named MODSM.
\begin{figure}[!t]
\centering
\subfloat{\includegraphics[width=1\linewidth]{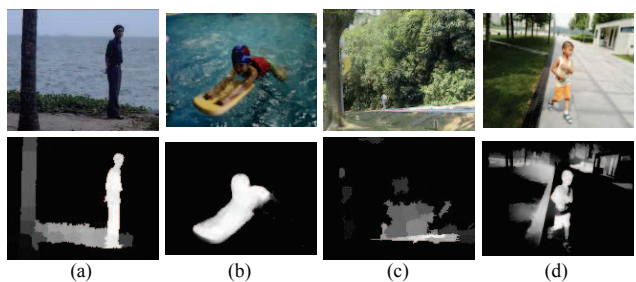}}
\caption{Images (top) and their saliency maps (bottom).}
\label{FIG.1}
\end{figure}
It is natural that ideal saliency map (e.g., the bottom of Fig. {\ref{FIG.1}}(a) and Fig. {\ref{FIG.1}}(b)) is desirable for the proposed method. However, even relatively unsatisfying saliency map (e.g., the bottom of Fig. {\ref{FIG.1}}(c) and Fig. {\ref{FIG.1}}(d)) can also play a positive role in the proposed MODSAM method. Of course, completely bad saliency map has a negative influence on moving object detection. Fortunately, great progress of salient object detection has been achieved \cite{Wang_cosegmentation_TIP_2015,Li_salient_TIP_2015} and their fruits can be borrowed for moving object detection.

Several methods were developed to employ a salient object detection algorithm for improving the performance of moving object detection \cite{Hao_spatiotemporal_CJA_2013}, \cite{Xia_spatiotemporal_IJCSI_2013}, \cite{Zhong_dynamic_CAI_2013}. Despite the initial success, their performance cannot arrive at the level of state-of-the-art low-rank based and subspace based methods \cite{Zhou_Low-rank_CoRR_2014}, \cite{Zhou_DECOLOR_PAMI_2013}, \cite{He_GRASTA_CVPR_2012}, \cite{Xu_GOSUS_ICCV_2013}, \cite{Guo_RFDSA_ECCV_2014}, \cite{Shakeri_COROLA_CoRR_2015}, \cite{Haines_Dirichlet_PAMI_2014}.

The rest of the paper is organized as follows. We review related work in Section II. The proposed method is given in Section III. Experimental results are provided in Section IV. We then conclude in Section V.

\section{Relater work}
Moving object detection can be implemented by different manners: detecting followed by tracking\cite{Pang_SVM_TCyber_2014,Dollar_pyramids_PAMI_2014, Isard_CONDENSATION_IJCV_1998}, subtracting frames\cite{Neri_Automatic_SP_1998,Wang_mobile_Neurocomputing_2013}, modeling background by density function, modeling background by subspace, modeling background by low-rank matrix. The last two manners dominate the state-of-the-art methods and are closely related to our work. Note that moving object detection methods can also be divided into incremental methods and batch methods. Our method belongs to incremental one.

\textbf{Subtracting frames} This kind of methods detects moving objects based on the differences between adjacent frames \cite{Neri_Automatic_SP_1998}, \cite{Wang_mobile_Neurocomputing_2013}. But these methods were proved not robust against illumination variations, changing background, camera motion, and noise.

\textbf{Modeling background by density function} This strategy assumes that the background is stationary and can be modeled by Gaussian, Mixture of Gaussians, or Dirichlet Process Mixture Models \cite{Haritaoglu_surveillance_PAMI_2000}, \cite{Li_Statistical_TIP_2004}, \cite{Haines_Dirichlet_PAMI_2014}. The foreground (moving objects) can then be obtained by subtracting the current frame with the background model.

\textbf{Modeling background by subspace} Instead of using a density function, subspace based method models the background as a linear combination the bases of a subspace \cite{He_GRASTA_CVPR_2012}, \cite{Xu_GOSUS_ICCV_2013}, \cite{Guo_RFDSA_ECCV_2014}, \cite{Yuan_IPCA_Neurocomputing_2009}, \cite{Pang_LDA_TNNLS_2014}. Because the subspace can be updated in an incremental (online) manner, its efficiency is very high. This kind of subspace based algorithms needs to impose constraints on the foreground in order to obtain valid solutions. Foreground sparsity is one of the widely used constraints which implies that the area of moving objects is small relative to the background. Principal Component Prusuit (PCP) \cite{Candes_RPCA_ACM_2011} is an important pioneer work which adopts   norm for measuring the foreground sparsity. It is the constraint of foreground sparsity that makes PCP suitable for foreground-background separation. Without this constraint, traditional robust subspace methods  can only deal with noise and outliers \cite{Favaro_RSEC_CVPR_2011}, \cite{Torre_RSL_IJCV_2003}, \cite{Qiu_link_arXiV_2011}, \cite{Vidal_GPCA_PAMI_2005}. The method \cite{Xin_BSGFLFM_CVPR_2015} improves PCP by taking into account the foreground connectivity (i.e., foreground structure). RFDSA takes into account smoothness and arbitrariness constraints \cite{Guo_RFDSA_ECCV_2014}.

But PCP \cite{Candes_RPCA_ACM_2011}, RFDSA\cite{Guo_RFDSA_ECCV_2014}, and the method \cite{Xin_BSGFLFM_CVPR_2015} are batch algorithms. Its detection speed cannot arrive at real-time level. Therefore, incremental (online) subspace methods are crucial for real-time detection \cite{Balzano_identification_PACC_2010}. He et al. \cite{He_GRASTA_CVPR_2012} proposed an online subspace tracking algorithm called GRASTA (Grassmannian Robust Adaptive Subspace Tracking Algorithm). Similar to PCP, GRASTA also explores   norm for imposing sparsity on foreground. But the GRASTA algorithm does not utilize any connectivity (a.k.a., smoothness) property of foreground. The GOSUS (Grassmannian Online Subspace Updates with Structured-sparsity) algorithm \cite{Xu_GOSUS_ICCV_2013} imposes a connectivity constraint on the objective function by grouping the pixels with a superpixel method and encouraging sparsity of the groups. Because of the large computational cost of the superpixel algorithm \cite{Achanta_SLIC_PAMI_2012}, GOSUS is not as efficient as GRASTA.

\textbf{Modeling background by low-rank matrix Low-rank modeling} is effective in video representation \cite{Zhou_Low-rank_CoRR_2014}. A sequence of vectorized images is represented as a matrix and the matrix is approximated by the sum of matrices of vectorized foreground, background, and noise \cite{Zhou_DECOLOR_PAMI_2013}. It is reasonable to assume that the background matrix is low-rank. DECOLOR (DEtecting Contiguous Outliers in the LOw-rank Representation) \cite{Zhou_DECOLOR_PAMI_2013} is considered as one of the most successful low-rank based algorithms. In DECOLOR, both foreground sparsity and contiguity (connectivity) are taken into account. It can be interpreted as  -penalty regularized RPCA \cite{Zhou_DECOLOR_PAMI_2013}. But the matrix computation can be started only if all of the predefined number of successive images is available. Obviously, such a batch method is not suitable for real-time video analysis due to its low efficiency. ISC \cite{Pang_IDSC_CC_2015} and COROLA \cite{Shakeri_COROLA_CoRR_2015} are incremental versions of DECOLOR. ISC and COROLA transforms low-rank method to subspace one.

The low-rank methods and subspace methods impose sparsity and connectivity (a.k.a., smoothness) on foreground and impose low-rank or principal components on background. In addition to such properties, in this paper we propose to impose saliency map on background and foreground meanwhile.
\section{Proposed method}
The proposed method belongs to incremental subspace based moving object 
detection method. Our main contribution lies in employing a saliency map to 
form a new objective function, resulting in fewer false and missed alarms.

\subsection{Input and Output}

The input of the Algorithm is a sequence of frames (images). Denote ${\rm 
{\bf o}} \in {\rm R}^{N\times 1}$ the current image and denote $o_i $ the 
$i$-th pixel of ${\rm {\bf o}}$. There are $N$ pixels in an image. The goal 
is to find the locations of the moving objects (i.e., foreground) in the 
current image ${\rm {\bf o}}$. The foreground locations are represented by a 
foreground-indicator vector ${\rm {\bf f}} \in \{0,1\}^N$. The $i$-element $f_i$ of $\bf f$ equals to either zero or one:

\begin{equation}
f_i  = \left\{ {\begin{array}{*{20}c}
   0 & {\text{if pixel i is classified as background,}}  \\
   1 & {\text{if pixel i is classified as foreground.}}  \\
\end{array}} \right.
\label{Eq.1}
\end{equation}
The foreground-indicator vector
${\rm\bf{f}}$ is obtained by binarizing background vector
${\rm{\bf b}} \in {\rm R}^{N\times 1}$ with a threshold ${\rm t}$:
\begin{equation}
f_i  = \left\{ {\begin{array}{*{20}c}
   0 & {{\text{if}}  ~b_i  \ge t,}  \\
   1 & {{\text{if}}   ~b_i  < t,}  \\
\end{array}} \right.
\label{Eq.2}
\end{equation}
where $b_i$ is the $i$-element of ${\bf{b}}$. The possibility of pixel $i$ being background increases with the value of $b_i$ and the possibility of pixel $i$ being foreground decreases with the increasing value of $b_i$.
\subsection{Problem Formulation}
As stated above (i.e., Eq. ({\ref{Eq.1}}) and Eq. ({\ref{Eq.2}}), the foreground-indicator vector can be obtained by binarizing background vector ${\bf{b}}$. The problem is how to compute ${\bf{b}}$ once a frame ${\bf{o}}$ is given. In this paper we formulate the problem of computing ${\bf{b}}$ as the following minimization problem:
\begin{equation}
\begin{split}
\mathop {\min }\limits_{{\bf{b}},{\bf{U}},{\bf{v}}} \sum\limits_{i = 1}^N \left[ {\frac{1}{2}b_i ({\bf{U}}_i {\bf{v}} - o_i )^2  + \beta (1 - b_i ) }\right.\\
\left.{- \alpha b_i (1 - s_i )} \right] +  \lambda \left\| {{\bf{Db}}} \right\|_1,
\end{split}
\label{Eq.3}
\end{equation}
where ${\bf{U}}_i$ stands for the $i$-th row of ${\bf{U}}$. In Eq. ({\ref{Eq.3}}),$s_i  \in [0,1]$ is the $i$-th element of the vector ${\rm{\bf s}} \in {\rm R}^{N\times 1}$ of a saliency map obtained by some salient object detection algorithm such as \cite{Zhu_RBD_CVPR_2014}. The value of $s_i $ reflects the confidence that the pixel $i$ belongs to a salient object. The term $- b_i (1 - s_i )$ is called saliency map term.

Minimizing the term $- b_i (1 - s_i )$ makes the estimated background ${\bf{b}}$ has less chance than foreground to contain salient objects. Moving objects such as pedestrian, car, dog are indeed salient objects in a video. Therefore, the proposed method is capable of making estimated foreground to have high-level semantic objects and fewer false alarms. The saliency map term $- b_i (1 - s_i )$ is the main novelty of the paper. $\alpha $ is the weight of the saliency map term. In Table {\ref{T.2}} , an empirical method for setting $\alpha $ is given.

In addition to $- b_i (1 - s_i )$, there are three terms: $b_i ({\bf{U}}_i {\bf{v}} - o_i )^2$, $(1 - s_i )$, and $\left\| {{\bf{Db}}} \right\|_1$ which are to be described as follows. The weights for $(1 - s_i )$ and $\left\| {{\bf{Db}}} \right\|_1$ are respectively $\beta $ and $\lambda $ whose values can be assigned according to Table {\ref{T.2}} .

The term $b_i ({\bf{U}}_i {\bf{v}} - o_i )^2$ is called background reconstruction term \cite{He_GRASTA_CVPR_2012}, \cite{Xu_GOSUS_ICCV_2013}. In this term, ${\rm{\bf U}} \in {\rm R}^{N\times m}$ is a subspace matrix and $m$ is the number of columns of ${\bf{U}}$. The vector ${\rm{\bf v}} \in {\rm R}^{N\times 1}$ is called coefficient vector. Minimizing the term $b_i ({\bf{U}}_i {\bf{v}} - o_i )^2$ makes the reconstructed background approaching the input frame ${\bf{o}}$ as possible as it can.

The term $(1 - s_i )$ is called foreground sparsity term. Minimizing $\beta (1 - s_i )$ makes the estimated foreground much sparser than background.

The term $\left\| {{\bf{Db}}} \right\|_1$ is called connectivity term \cite{Xu_GOSUS_ICCV_2013}, \cite{Guo_RFDSA_ECCV_2014}. The matrix ${\rm{\bf D}} \in {\rm R}^{2N\times N}$ is a difference matrix \cite{Xu_GOSUS_ICCV_2013}, \cite{Guo_RFDSA_ECCV_2014}. Minimizing $\left\| {{\bf{Db}}} \right\|_1$ makes estimated background and foreground smooth as possible as it can. That is, if a pixel belongs to background (or foreground), then its neighbors also belong to background (or foreground).
\subsection{Problem Solution}
To obtain the solution to Eq. ({\ref{Eq.3}}), Eq. ({\ref{Eq.3}}) is equivalently transformed to the following problem:
\begin{equation}
\begin{split}
\mathop {\min }\limits_{{\bf{b}},{\bf{U}},{\bf{v}}} \sum\limits_{i = 1}^N \left[ {\frac{1}{2}b_i ({\bf{U}}_i {\bf{v}} - o_i )^2  + \beta (1 - b_i )}\right. \\ 
\left.{- \alpha b_i (1 - s_i )} \right] +  \lambda \left\| {\bf{c}} \right\|_1,
\end{split}
\label{Eq.4}
\end{equation}
\begin{equation}
s.t.{\rm{ }}{\bf{c}} = {\bf{Dw}},{\rm{ }}{\bf{w}} = {\bf{b}}.
\label{Eq.5}
\end{equation}

The constrained minimum problem expressed as Eq. ({\ref{Eq.4}}) and Eq. ({\ref{Eq.5}}) can be converted to the following unconstrained problem:
\begin{equation}
\begin{split}
\mathop {\min }\limits_{{\bf{b}},{\bf{U}},{\bf{v,c,w,x,y}}} \sum\limits_{i = 1}^N  \left[ {\frac{1}{2}b_i ({\bf{U}}_i {\bf{v}} - o_i )^2  + \beta (1 - b_i ) }\right.\\
\left.{- \alpha b_i (1 - s_i )} \right] + \lambda \left\| {\bf{c}} \right\|_1  + \frac{\mu }{2}\left\| {{\bf{w - b}}} \right\|_F^2 \\
+ {\bf{x}}^T ({\bf{w - b}})+ \frac{\mu }{2}\left\| {{\bf{c - Dw}}} \right\|_F^2 + {\bf{y}}^T ({\bf{c - Dw}}).
\end{split}
\label{Eq.6}
\end{equation}

We adopt an alternative minimization algorithm to seek the optimal solutions of Eq. ({\ref{Eq.6}}). The steps are as follows.

${\bf{b-Step}}$. The goal is to seek the optimal ${\bf{b}}$ when ${\bf{U}}$, ${\bf{v}}$, ${\bf{c}}$, ${\bf{w}}$, ${\bf{x}}$, and ${\bf{y}}$ are fixed. Computing the derivative of the sum of the terms of Eq. ({\ref{Eq.6}}) and letting the result be zero yields
\begin{equation}
b_i  = \frac{{\beta  + \mu w_i  + x_i  - \frac{1}{2}({\bf{U}}_i {\bf{v}} - o_i )^2  + \alpha (1 - s_i )}}{\mu }.
\label{Eq.7}
\end{equation}

The influence of the saliency map $s_i$ on the background $b_i$ is intuitive: $b_i$ decreases with increasing of $s_i$. Hence, the proposed method tends to let estimated background not contain moving and salient objects whereas let estimated foreground contain moving and salient objects meanwhile.

${\bf{c-Step}}$. The goal is to seek the optimal ${\bf{c}}$ when ${\bf{b}}$, ${\bf{U}}$, ${\bf{v}}$, ${\bf{w}}$, ${\bf{x}}$, and ${\bf{y}}$ are fixed. Omitting irrelevant terms, it is reduced to the following traditional optimization problem:
\begin{equation}
{\bf{c}} = \arg \mathop {\min }\limits_{\bf{c}} {\rm{ }}\lambda \left\| {\bf{c}} \right\|_1  + \frac{\mu }{2}\left\| {{\bf{c - Dw}}} \right\|_F^2  + {\bf{y}}^T ({\bf{c - Dw}})
\label{Eq.8}
\end{equation}
\begin{equation}
{\rm{ = arg}}\mathop {{\rm{min}}}\limits_{\bf{c}} {\rm{ }}\frac{\lambda }{\mu }\left\| {\bf{c}} \right\|_1  + \frac{1}{2}\left\| {{\bf{c - m}}} \right\|_F^2,
\label{Eq.9}
\end{equation}
where
\begin{equation}
{\bf{m}} = {\bf{Dw}} - \frac{{\bf{y}}}{\mu }.
\label{Eq.10}
\end{equation}
Eq. ({\ref{Eq.9}})  is standard minimization problem \cite{Boyd_Distributed_FTML_2011} and the solution is given by \cite{Guo_RFDSA_ECCV_2014}
\begin{equation}
{\bf{c}} = S_{\frac{\lambda }{\mu }} ({\bf{Dw}} - \frac{{\bf{y}}}{\mu })
\label{Eq.11}
\end{equation}
with the soft-thresholding (shrinkage) operator $S_\varepsilon  (x)$ being
\begin{small}
\begin{equation}
S_\varepsilon  (x) = {\mathop{\rm sgn}} (x)\max (|x| - \varepsilon ,0){\rm{ = }}\left\{ {\begin{array}{*{20}c}
   {x - \varepsilon ,} \hfill & {x > \varepsilon } \hfill  \\
   {x + \varepsilon ,} \hfill & {x <  - \varepsilon } \hfill  \\
   0 \hfill & {else} \hfill  \\
\end{array}} \right.
\label{Eq.12}
\end{equation}
\end{small}

${\bf{w-Step}}$. The goal is to seek the optimal ${\bf{w}}$ when ${\bf{b}}$, ${\bf{U}}$, ${\bf{v}}$, ${\bf{c}}$, ${\bf{x}}$, and ${\bf{y}}$ are fixed. Omitting irrelevant terms, it is reduced to the following minimization problem:
\begin{equation}
\begin{split}
{\bf{w}}{\rm{ = }}\arg \mathop {\min }\limits_{\bf{w}} {\rm{ }}\frac{\mu }{2}\left\| {{\bf{w - b}}} \right\|_F^2  + {\bf{x}}^T ({\bf{w - b}}) \\ 
+ \frac{\mu }{2}\left\| {{\bf{c - Dw}}} \right\|_F^2 + {\bf{y}}^T ({\bf{c - Dw}}).
\end{split}
\label{Eq.13}
\end{equation}
Specifically, the optimal ${\bf{w}}$ is calculated by
\begin{equation}
{\bf{w}} = ({\bf{I + D}}^T {\bf{D}})^{ - 1} \left[ {{\bf{D}}^T ({\bf{c}} + \frac{{\bf{y}}}{\mu }) + {\bf{b}} - \frac{{\bf{x}}}{\mu }} \right].
\label{Eq.14}
\end{equation}

${\bf{x,y-Step}}$. The goal is to seek the optimal ${\bf{x}}$ and ${\bf{y}}$ when ${\bf{b}}$, ${\bf{U}}$, ${\bf{v}}$, ${\bf{c}}$, and ${\bf{w}}$ are fixed. Computing the derivative of the sum of the terms of Eq. ({\ref{Eq.6}}) w.r.t. ${\bf{x}}$ and ${\bf{y}}$ and then letting the result be zero yields the following updating rule:
\begin{equation}
{\bf{x}} \leftarrow {\bf{x}} + \mu ({\bf{w}} - {\bf{b}}),
\label{Eq.15}
\end{equation}
\begin{equation}
{\bf{y}} \leftarrow {\bf{y}} + \mu ({\bf{c - Dw}}).
\label{Eq.16}
\end{equation}
It is noted that the coefficient $\mu $ is updated by
\begin{equation}
\mu  \leftarrow a\mu,
\label{Eq.17}
\end{equation}
where $a$ is a parameter and its empirical value is 1.25.

${\bf{U-Step}}$. The goal is to seek the optimal ${\bf{U}}$ when ${\bf{b}}$, ${\bf{v}}$, ${\bf{c}}$, ${\bf{w}}$, ${\bf{x}}$, and ${\bf{y}}$ are fixed. The problem of minimizing Eq. ({\ref{Eq.6}}) with respect to ${\bf{U}}$ becomes
\begin{equation}
{\bf{U}} = \arg \mathop {\min }\limits_{\bf{U}} {\rm{ }}\sum\limits_i {\frac{1}{2}b_i ({\bf{U}}_i {\bf{v}} - o_i )^2 } ,{\rm{ }}s.t.{\rm{ }}{\bf{UU}}^T  = {\bf{I}},
\label{Eq.18}
\end{equation}
where ${\bf{I}}$ is the identity matrix. It is known that orthogonal matrices representing linear subspaces of the Euclidean space can be represented as points on the Grassmann manifolds \cite{Mittal_Grassmann_IVC_2012}. So subspace estimation can be equivalently formulated into an optimization problem on Grassmann manifolds \cite{Mittal_Grassmann_IVC_2012}. Defining
\begin{equation}
L_f  \buildrel \Delta \over = \frac{1}{2}{\bf{b}}({\bf{UV - o)v}}^T {\bf{U}} = \sum\limits_i {\frac{1}{2}b_i ({\bf{U}}_i {\bf{v}} - o_i )^2 },
\label{Eq.19}
\end{equation}
the optimization can be performed by using the gradient $\partial L_f /\partial {\bf{U}}$ on the Euclidean space and the gradient $\nabla L_f $ of the Grassmannian \cite{Edelman_orthogonality_MAA_1998}. The gradients are given by
\begin{equation}
\frac{{\partial L_f }}{{\partial {\bf{U}}}} = {\bf{b}}({\bf{UV - o)v}}^T,
\label{Eq.20}
\end{equation}
and
\begin{equation}
\begin{array}{l}
 \nabla L_f  = ({\bf{I}} - {\bf{UU}}^T )\frac{{\partial L_f }}{{\partial {\bf{U}}}} \\ 
 {\rm{        = }}({\bf{I}} - {\bf{UU}}^T ){\bf{b}}({\bf{UV - o)v}}^T  \\ 
 {\rm{        = }}({\bf{I}} - {\bf{UU}}^T ){\bf{Rv}}^T  \\ 
 \end{array},
 \label{Eq.21}
\end{equation}
where the residual vector ${\bf{R}}$ is defined as
\begin{equation}
{\rm{ }}{\bf{R}} \buildrel \Delta \over = {\bf{b}}({\bf{Uv - o)}}.
\label{Eq.22}
\end{equation}
The solution on the Grassmannian manifolds is \cite{He_GRASTA_CVPR_2012}, \cite{Xu_GOSUS_ICCV_2013}.
\begin{equation}
\begin{split}
{\bf{U}} \leftarrow  = {\bf{U}} + (\cos (\sigma \eta ) - 1){\bf{U}}\frac{{\bf{v}}}{{\left\| {\bf{v}} \right\|}}\frac{{{\bf{v}}^T }}{{\left\| {\bf{v}} \right\|}} \\ 
- \sin (\sigma \eta )\frac{{\bf{R}}}{{\left\| {\bf{R}} \right\|}}\frac{{\bf{v}}}{{\left\| {\bf{v}} \right\|}}.
\end{split}
\label{Eq.23}
\end{equation}

${\bf{v-Step}}$. The low-dimensional representation ${\bf{v}}$ of ${\bf{o}}$ can be simply calculated by
\begin{equation}
{\bf{v = U}}^T {\bf{o}}.
\label{Eq.24}
\end{equation}

\begin{algorithm}[!t]
\renewcommand{\algorithmicrequire}{\textbf{Input:}}
\renewcommand\algorithmicensure {\textbf{Output:} }
\renewcommand\algorithmicrepeat {\textbf{Iteration} }
\caption{The proposed method of moving object detection.}
\begin{algorithmic}[1]
\REQUIRE ~~\\
A sequence of frames (images) and the current image is ${\bf o}$. Each image has $N$ pixels. 
\ENSURE ~~\\ 
Foreground-indicator vector ${\bf f}$ corresponding to the current image ${\bf o}$. 
\STATE \textbf{Initialization}

\STATE Initialize parameters $\alpha$, $\beta$, $\mu$, $\lambda $.

\STATE Initialize ${\bf U}$, ${\bf v}$, ${\bf c}$, ${\bf w}$, ${\bf x}$, and ${\bf y}$. 

\STATE Applying some salient object diction algorithm on ${\bf o}$ and get the corresponding saliency map ${\bf s}$.

\STATE Iterating the following steps several loops

\STATE Begin Iteration: 

\STATE ${\bf b-Step}$:$ b_i  = \frac{{\beta  + \mu w_i  + x_i  - \frac{1}{2}({\bf{U}}_i {\bf{v}} - o_i )^2  + \alpha (1 - s_i )}}{\mu } $

\STATE ${\bf c-Step}$:$ {\bf{c}} = S_{\frac{\lambda }{\mu }} ({\bf{Dw}} - \frac{{\bf{y}}}{\mu }) $

\STATE ${\bf w-Step}$:$ {\bf{w}} = ({\bf{I + D}}^T {\bf{D}})^{ - 1} \left[ {{\bf{D}}^T ({\bf{c}} + \frac{{\bf{y}}}{\mu }) + {\bf{b}} - \frac{{\bf{x}}}{\mu }} \right] $

\STATE ${\bf x,y-Step}$:Assign a small number to $\mu$ . Update ${\bf x}$ and ${\bf y}$ by running the following formulas several loops:
$ {\bf{x}} \leftarrow {\bf{x}} + \mu ({\bf{w}} - {\bf{b}}), {\bf{y}} \leftarrow {\bf{y}} + \mu ({\bf{c - Dw}}), \mu  \leftarrow 1.25\mu. $

\STATE ${\bf U-Step}$:Assign a small number to $\eta$. Update ${\bf U}$ by running the following formulas several loops:
$ {\bf{R}}{\rm{ = }}{\bf{b}}({\bf{Uv - o)}},
{\bf{U}} \leftarrow  = {\bf{U}} + (\cos (\sigma \eta ) - 1){\bf{U}}\frac{{\bf{v}}}{{\left\| {\bf{v}} \right\|}}\frac{{{\bf{v}}^T }}{{\left\| {\bf{v}} \right\|}} - \sin (\sigma \eta )\frac{{\bf{R}}}{{\left\| {\bf{R}} \right\|}}\frac{{\bf{v}}}{{\left\| {\bf{v}} \right\|}}. $

\STATE End Iteration

\STATE Compute foreground-indicator vector ${\bf f}$ is obtained by binarizing background vector:$ f_i  = \left\{ {\begin{array}{*{20}c}
   0 & {\text{if} ~b_i  \ge t,}  \\
   1 & {\text{if}  ~b_i  < t.}  \\
\end{array}} \right. $

\end{algorithmic}
\end{algorithm}

Algorithm 1 summarizes the above steps.

\section{Experimental results}
We describe intermediate results followed by comparison with state-of-the-art methods. In our experiments, the saliency maps are obtained by the method developed in \cite{Zhu_RBD_CVPR_2014}.
\subsection{Intermediate results}
We give intermediate results to show the role of the saliency map term $- b_i (1 - s_i )$ and the connectivity term $\left\| {{\bf{Db}}} \right\|_1$.

For notation simplicity, in Table {\ref{T.1}}  we list the objective functions of three methods: Baseline, Add Connectivity, and Add Saliency Map. The objective function of the proposed method is
\begin{equation}
\begin{split}
L = \mathop {\min }\limits_{{\bf{b}},{\bf{U}},{\bf{v}}} \sum\limits_{i = 1}^N \left[ {\frac{1}{2}b_i ({\bf{U}}_i {\bf{v}} - o_i )^2  + \beta (1 - b_i )}\right.\\
 \left.{- \alpha b_i (1 - s_i )} \right] +  \lambda \left\| {{\bf{Db}}} \right\|_1.
\end{split}
\label{Eq.25}
\end{equation}
The baseline is the method whose objective function $L_b$ (Eq. ({\ref{Eq.26}})) consists of the first two terms of $L$ (Eq. ({\ref{Eq.25}})):
\begin{equation}
\begin{split}
L_b  = \mathop {\min }\limits_{{\bf{b}},{\bf{U}},{\bf{v}}} \sum\limits_{i = 1}^N {\left[ {\frac{1}{2}b_i ({\bf{U}}_i {\bf{v}} - o_i )^2  + \beta (1 - b_i )} \right]}.
\end{split}
\label{Eq.26}
\end{equation}
In addition to the reconstruction term, the baseline method merely makes use of the sparsity term $\beta (1 - b_i )$.

Compared to $L_b$ (Eq. ({\ref{Eq.26}})), the objective function $L_c$ (Eq. ({\ref{Eq.27}})) of Add Connectivity has additional connectivity term $\lambda \left\| {{\bf{Db}}} \right\|_1$:
\begin{equation}
\begin{split}
L_c  = \mathop {\min }\limits_{{\bf{b}},{\bf{U}},{\bf{v}}} \sum\limits_{i = 1}^N {\left[ {\frac{1}{2}b_i ({\bf{U}}_i {\bf{v}} - o_i )^2  + \beta (1 - b_i )} \right]} \\
 +  \lambda \left\| {{\bf{Db}}} \right\|_1.
\end{split}
\label{Eq.27}
\end{equation}

The objective function of Add Saliency Map is the same as $L$ (Eq. ({\ref{Eq.25}})). That is, Add Saliency Map is the final form of our method where sparsity, low-rank, connectivity, and saliency map are taken into account.

\begin{table}[!t]
\renewcommand{\arraystretch}{1.4}
\caption{Method used for intermediate results.}
\centering
\begin{tiny}
\begin{tabular}{p{1cm}p{7.5cm}}
\hline
Method & Objective Function\\
\hline
Baseline & $L_b  = \mathop {\min }\limits_{{\bf{b}},{\bf{U}},{\bf{v}}} \sum\limits_{i = 1}^N {\left[ {\frac{1}{2}b_i ({\bf{U}}_i {\bf{v}} - o_i )^2  + \beta (1 - b_i )} \right]}$ \\
Add Connectivity & $ L_c  = \mathop {\min }\limits_{{\bf{b}},{\bf{U}},{\bf{v}}} \sum\limits_{i = 1}^N {\left[ {\frac{1}{2}b_i ({\bf{U}}_i {\bf{v}} - o_i )^2  + \beta (1 - b_i )} \right] + } \lambda \left\| {{\bf{Db}}} \right\|_1 $ \\
Add Saliency Map &  $ L = \mathop {\min }\limits_{{\bf{b}},{\bf{U}},{\bf{v}}} \sum\limits_{i = 1}^N {\left[ {\frac{1}{2}b_i ({\bf{U}}_i {\bf{v}} - o_i )^2  + \beta (1 - b_i ) - \alpha b_i (1 - s_i )} \right]} +  \lambda \left\| {{\bf{Db}}} \right\|_1 $  \\
\hline
\end{tabular}
\end{tiny}
\label{T.1}
\end{table}

\begin{figure}[!t]
\centering
\subfloat{\includegraphics[width=1\linewidth]{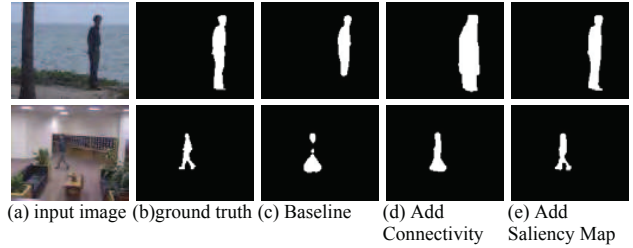}}
\caption{The influence of adding connectivity and saliency map to the objective function.}
\label{fig.2}
\end{figure}

\begin{figure}[!t]
\centering
\subfloat{\includegraphics[width=1\linewidth]{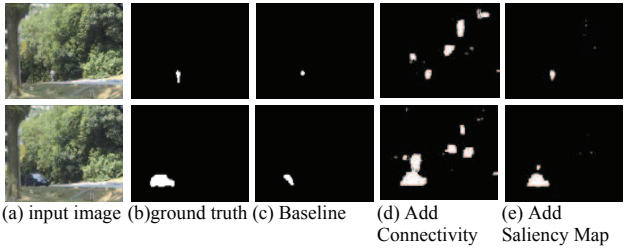}}
\caption{The influence of adding connectivity and saliency map to the objective function.}
\label{fig.3}
\end{figure}

Several frames of the Perception Test Image Sequences \cite{Li_Statistical_TIP_2004} are used for analyzing the intermediate results. Some examples are shown in Fig. {\ref{fig.2}} and Fig. {\ref{fig.3}}. Fig. {\ref{fig.2}} (a) shows two input frames with water-surface background for the top one and indoor environment for the bottom one. The ground truth of the moving object is given in Fig. {\ref{fig.2}} (b). Fig. {\ref{fig.2}} (c) is the detected results of the baseline from which one can see that the detected object is smaller than the ground truth. The top of Fig. {\ref{fig.2}} (c) shows that the feet and some portions of the shanks are missed by the baseline method. The bottom of Fig. {\ref{fig.2}} (c) shows that the middle of the person is missed by the baseline method.

As can be seen from Fig. {\ref{fig.2}} (d), with the help of connectivity term, the Add Connectivity is able to detect the missed parts (feet and legs in the top of Fig. {\ref{fig.2}} (d) and the middle part of the person in the bottom of Fig. {\ref{fig.2}} (d)) of the persons. But one can also there are many false alarms in Fig. {\ref{fig.2}} (d). False alarms are the by-product of Add Connectivity. Fig. {\ref{fig.2}} (e) is the result of Add Saliency Map. Obviously, introducing the saliency map successfully discards the false alarms existing in Fig. {\ref{fig.2}} (d).

The results given in Fig. {\ref{fig.3}} (e) demonstrate that adding saliency map into the objective function is capable of suppressing many false alarms when the size of moving objects (a person in the top of Fig. {\ref{fig.3}} (a) and a car in the bottom of Fig. {\ref{fig.3}} (a)) is small whereas the background is large, complex and dynamic. Fig. {\ref{fig.3}} (d) shows that adding connectivity into the objective function not only enlarges the objects detected by the baseline but also incorrectly classifies moving leafs and shadows as semantic objects. Adding saliency map (Fig. {\ref{fig.3}} (e)) plays a role of overcoming the drawback of adding connectivity.

\begin{figure}[!t]
\centering
\subfloat{\includegraphics[width=1\linewidth]{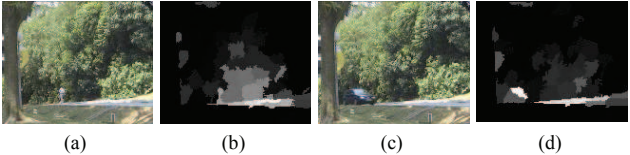}}
\caption{(b) and (c) are the saliency maps of (a) and (c), respectively.}
\label{fig.4}
\end{figure}

The saliency maps of the top and bottom of Fig. {\ref{fig.3}} (a) are shown in Fig. {\ref{fig.4}} (b) and Fig. {\ref{fig.4}} (d), respectively. Though the saliency maps are not ideal, they provide useful clue for the proposed method (i.e., Add Saliency Map).

\begin{figure}[!t]
\centering
\subfloat{\includegraphics[width=1\linewidth]{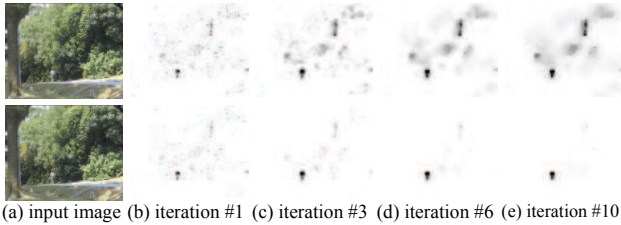}}
\caption{Input image and the background vector   obtained in different iteration of the Add Connectivity (Top) and Add Saliency Map (bottom).}
\label{fig.5}
\end{figure}

\begin{figure}[!t]
\centering
\subfloat{\includegraphics[width=1\linewidth]{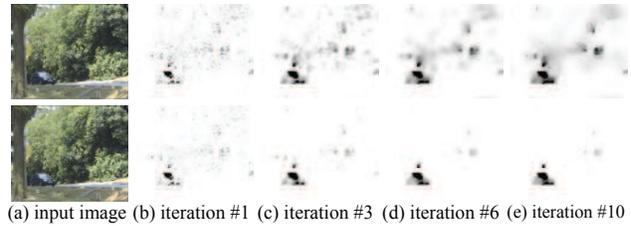}}%
\caption{Input image and the background vector $\bf b$  obtained in different iteration of the Add Connectivity (Top) and Add Saliency Map (bottom).}
\label{fig.6}
\end{figure}

The proposed Add Saliency Map algorithm (see Algorithm 1) and Add Connectivity algorithm update the background vector $\bf b$ iteratively. Fig. {\ref{fig.5}} and Fig. {\ref{fig.6}} show how the background vector $\bf b$ varies with iterations. Fig. {\ref{fig.5}} (a) shows the input image identical to the top of Fig. {\ref{fig.3}} (a). The top and bottom of Fig. {\ref{fig.5}} corresponds to the iteration results of Add Connectivity algorithm and Add Saliency Map algorithm, respectively. One can see from the top of Fig. {\ref{fig.5}} that the background vector obtained by Add Connectivity contains more regions of waving leafs as iteration proceeds. But one can see from the bottom of Fig. {\ref{fig.5}} that the background vector obtained by Add Saliency Map excludes more regions of waving leafs as iteration proceeds and hence the foreground vector focuses on the true meaningful moving person.

Similar to Fig. {\ref{fig.5}}, the bottom of Fig. {\ref{fig.6}} also demonstrates that adding the saliency map into the objective function makes the estimated background vector iteratively excludes the influence of moving leafs.

\begin{figure}[!hbp]
\centering
\subfloat[]{\includegraphics[width=0.5\linewidth]{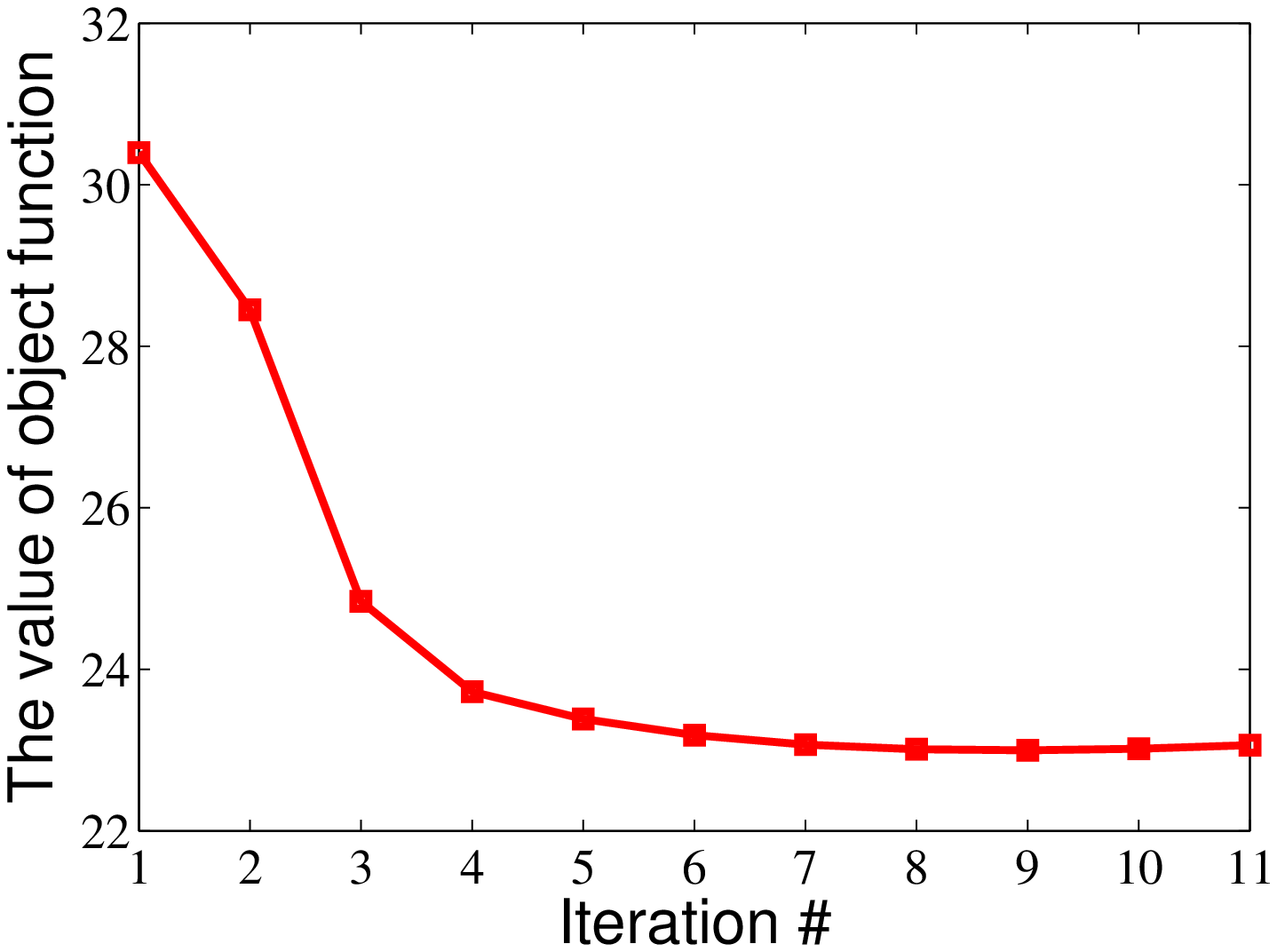}}
\subfloat[]{\includegraphics[width=0.5\linewidth]{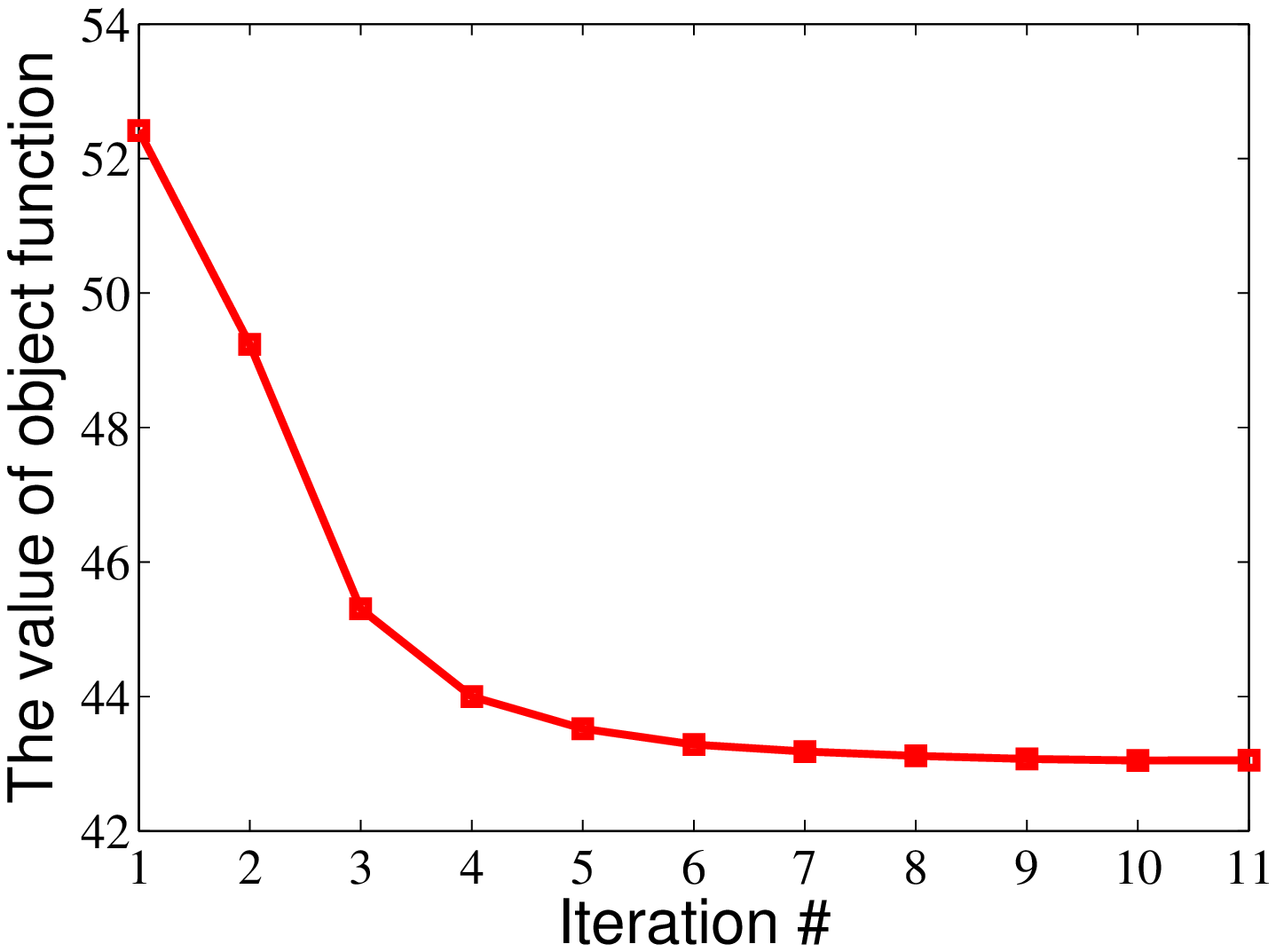}}
\caption{Convergence of the Add Saliency Map. (a) For the input image shown in Fig. {\ref{fig.5}}(a). (b) For the input image shown in Fig. {\ref{fig.6}}(a).}
\label{fig.7}
\end{figure}

Fig. {\ref{fig.7}}  shows that the convergence of the proposed algorithm. Generally, the value of the objective function $L$ decreases drastically at the first five iterations and becomes stable after iteration \# 8.

\subsection{Comparison with the State-of-the-art methods}
We call the proposed method Moving Object Detection using Saliency Map with abbreviation MODSM.

The Perception Test Images Sequences \cite{Li_Statistical_TIP_2004} are also used for comparison with the state-of-the-art methods. The dataset consists of 9 videos captured in a variety of indoor and outdoor environments, including offices, campuses, sidewalks, and other private and public sites.

The weather conditions when collecting the data cover sunny, cloudy, and rainy weather. The videos with static background are named Bootstrap (BS), Shopping Mall (SM), and Hall (Hal). The videos with dynamic background are called Fountain (Fou), Escalator (Esc), Water Surface (WS), Curtain (Cur), and Campus (Cam). The Lobby (Lob) video is captured when there are drastic variations in illumination. The sizes (widths and heights) of the frames includes [160, 130], [160, 128], [176, 144], [160, 120], [160, 128], and [320, 256].

We compare the proposed MODSM algorithm with PCP (Principal Component Prusuit) \cite{Candes_RPCA_ACM_2011}, DP-GMM (Dirichlet Process – Gaussian Mixture Models) \cite{Haines_Dirichlet_PAMI_2014}, GMM \cite{Stauffer_mixture_ICCV_1999}, GRASTA (Grassmannian Robust Adaptive Subspace Tracking Algorithm) \cite{He_GRASTA_CVPR_2012}, DECOLOR (DEtecting Contiguous Outliers in the LOw-rank Representation) \cite{Zhou_DECOLOR_PAMI_2013}, SOBS \cite{Maddalena_surveillance_TIP_2008} and RFDSA \cite{Guo_RFDSA_ECCV_2014}. PCP, GRASTA, and RFDSA are the state-of-the-art subspace based algorithms. DP-GMM is the state-of-the-art density based algorithm and DECOLOR is the state-of-the-art low-rank based algorithm. DP-GMM and GRASTA, are incremental algorithms whereas PCP, DECOLOR, and RFDSA are batch algorithms. We run the source codes provided by the authors of four methods on the dataset to get the experimental results. Note that GRASTA randomly samples a fraction of pixels in an image for subspace modeling and object detecting. Its detection accuracy increases with the fraction. To reduce randomness and get its best accuracy, 100\% pixels are used in our experiments.

The parameters (see Eq. ({\ref{Eq.6}})) of the MODSM method are given in Table {\ref{T.2}} where $m$ is the number of columns (basis vectors) of the matrix $\bf U$. In Table {\ref{T.2}} , $\hat \sigma ^2$ is given by
\begin{equation}
\hat \sigma ^2  = \frac{1}{{{\rm{2|}}\Omega {\rm{|}}}}\sum\limits_{{\bf{O}} \in \Omega } {{\rm{||}}{\bf{Uv - o}}{\rm{||}}_2^2 },
\label{Eq.28}
\end{equation}
where $\Omega$ and $|\Omega |$ are the set and the number of training images, respectively. $s_m$ is the ratio of the number of pixels whose saliency are larger than the mean of saliency maps of training images:
\begin{equation}
s_m  = \frac{{\sum\limits_{\bf{S}} {\sum\limits_{i = 1}^N {I(s_i  - s_M )} } }}{{N{\rm{|}}\Omega {\rm{|}}}},
\label{Eq.29}
\end{equation}
with
\begin{equation}
s_M  = \frac{{\sum\limits_{\bf{S}} {\sum\limits_{i = 1}^N {s_i } } }}{{N{\rm{|}}\Omega {\rm{|}}}},
\label{Eq.30}
\end{equation}
and
\begin{equation}
I(x) = \left\{ {\begin{array}{*{20}c}
   1 & {x > 0}  \\
   0 & {x \le 0}  \\
\end{array}} \right.
\label{Eq.31}
\end{equation}
Note that the $\left\lceil x \right\rceil $ in Table {\ref{T.2}} stands for the floor function of $x$.

Table {\ref{T.2}} gives a general rule for parameter setting. But the detection performance can be significantly improved if video-specific parameters are utilized.

\begin{table}[!hbp]
\renewcommand{\arraystretch}{1.5}
\caption{Parameters of the MODSM method.}
\centering
\begin{tiny}
\begin{tabular}{c c c c c}
\hline
$m$ & $\beta$ & $\lambda$ & $\alpha$ & $\mu$ \\
\hline
5 & $ \beta  = \max (\frac{1}{2}\beta ,4.5\hat \sigma ^2 )$ & $5\beta$ & $ \min \left( {\left\lceil {\frac{{s_m }}{{s_m  - s_M }}} \right\rceil \hat \sigma s_m ,6.5\beta } \right) $ & $0.1$ \\
\hline
\end{tabular}
\end{tiny}
\label{T.2}
\end{table}

The $F_1$-score, the harmonic mean of precision and recall, is used for objective evaluation:
\begin{equation}
F_1  = 2\frac{{precision \times recall}}{{precision + recall}}.
\label{Eq.32}
\end{equation}

\begin{table}[htbp]
\renewcommand{\arraystretch}{1.5}
\caption{F1-scores of different methods.}
\centering
\begin{tiny}
\begin{tabular}
{|p{21pt}|p{9pt}|p{9pt}|p{9pt}|p{9pt}|p{9pt}|p{9pt}|p{9pt}|p{9pt}|p{9pt}|p{9pt}|}
\hline
\textit{Method}& 
\textit{WS}& 
\textit{Cur}& 
\textit{Fou}& 
\textit{Hal}& 
\textit{SM}& 
\textit{Lob}& 
\textit{Esc}& 
\textit{BS}& 
\textit{Cam}& 
\textit{mean} \\
\hline
GMM& 
.7948& 
.7580& 
.6854& 
.3335& 
.5363& 
.6519& 
.1388& 
.3838& 
.0757& 
.4842 \\
\hline
SOBS& 
.8247& 
.8178& 
.6554& 
.5943& 
.6677& 
.6489& 
.5770& 
.6019& 
.6960& 
.6760 \\
\hline
DP-GMM& 
.9090& 
.8203& 
.7049& 
.5484& 
.6522& 
.5794& 
.5055& 
.6024& 
.7567& 
.6754 \\
\hline
PCP& 
.4137& 
.6193& 
.5679& 
.5917& 
.7234& 
.6989& 
.6728& 
.6582& 
.3406& 
.5874 \\
\hline
DECOLOR& 
.8866& 
.8255& 
.\textbf{8598}& 
.6424& 
.6525& 
.6149& 
.6994& 
.5869& 
.\textbf{8096}& 
.7308 \\
\hline
GRASTRA& 
.7310& 
.6591& 
.3786& 
.5817& 
.7142& 
.5550& 
.4697& 
.6146& 
.2504& 
.5505 \\
\hline
RFDSA& 
.8796& 
.8976& 
.7544& 
.6673& 
.\textbf{7407}& 
.\textbf{8029}& 
.6353& 
.6841& 
.6779& 
.7489 \\
\hline
\textbf{MODSM}& 
\textbf{.9404}& 
.\textbf{9098}& 
.8205& 
.\textbf{6859}& 
.7362& 
.5762& 
.\textbf{7553}& 
.\textbf{7280}& 
.7876& 
.\textbf{7711} \\
\hline
\end{tabular}
\end{tiny}
\label{T.3}
\end{table}

The results of the different methods are given in Table {\ref{T.3}}. Among the nine videos, the proposed MODSM, RFDSA, and DECOLOR get the best performance on five (i.e.,WS, Cur, Hal, Esc, and BS ), two (i.e., SM and Lob), and two (Fou and Cam) different videos, respectively. The average $F_1$-score of the proposed MODSM is the largest. But our method does not work well for the Lobby (i.e., Lob) video. The main reason is that the performance of the method \cite{Zhu_RBD_CVPR_2014} of creating saliency map on the Lobby video degraded significantly. If the Lobby video is excluded, the average $F_1$-score of MODSM grows from 0.7711 to 0.7955 whereas that of RFDSA decreases from 0.7489 to 0.7421. It is expected that the performance of MODSM increases with the performance of saliency map.

Table {\ref{T.3}} also shows that if proper prior information (i.e., connectivity, saliency map, sparsity) is employed then the incremental algorithm MODSM can outperform the batch algorithms DECOLOR and RFDSA.

\begin{figure}[!hbp]
\centering
\subfloat[]{\includegraphics[width=0.5\linewidth]{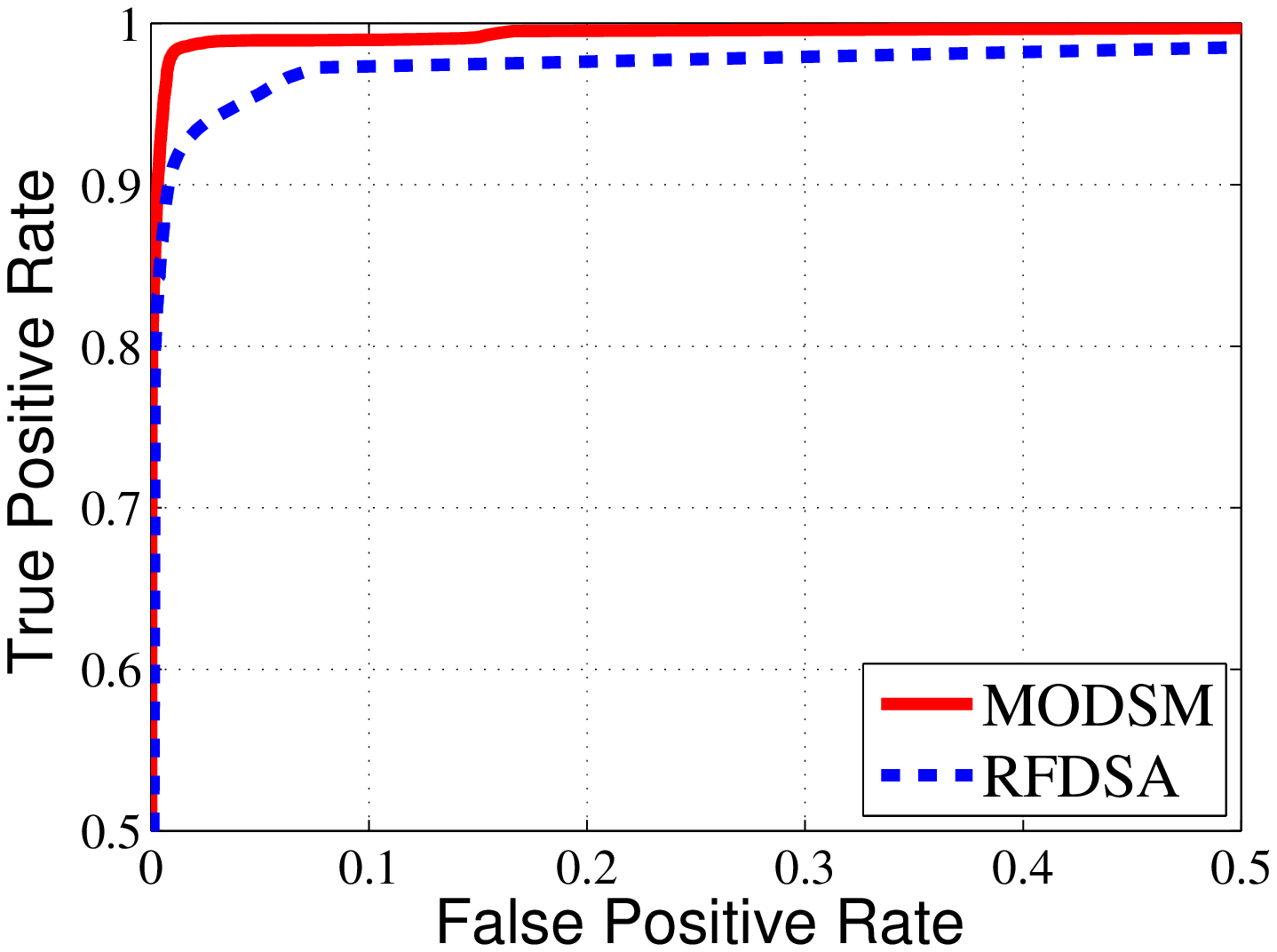}}
\subfloat[]{\includegraphics[width=0.5\linewidth]{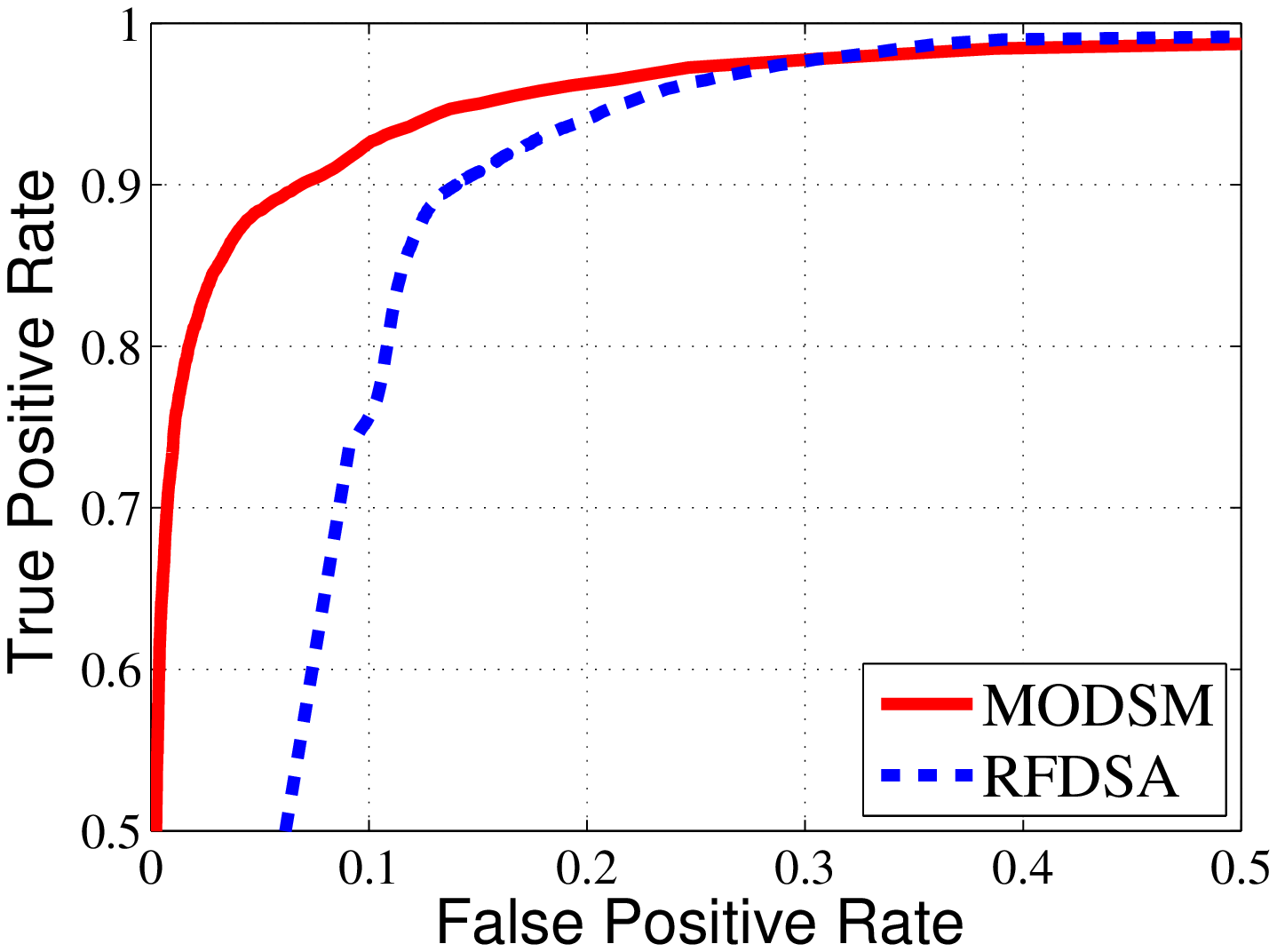}} \\ 
\subfloat[]{\includegraphics[width=0.5\linewidth]{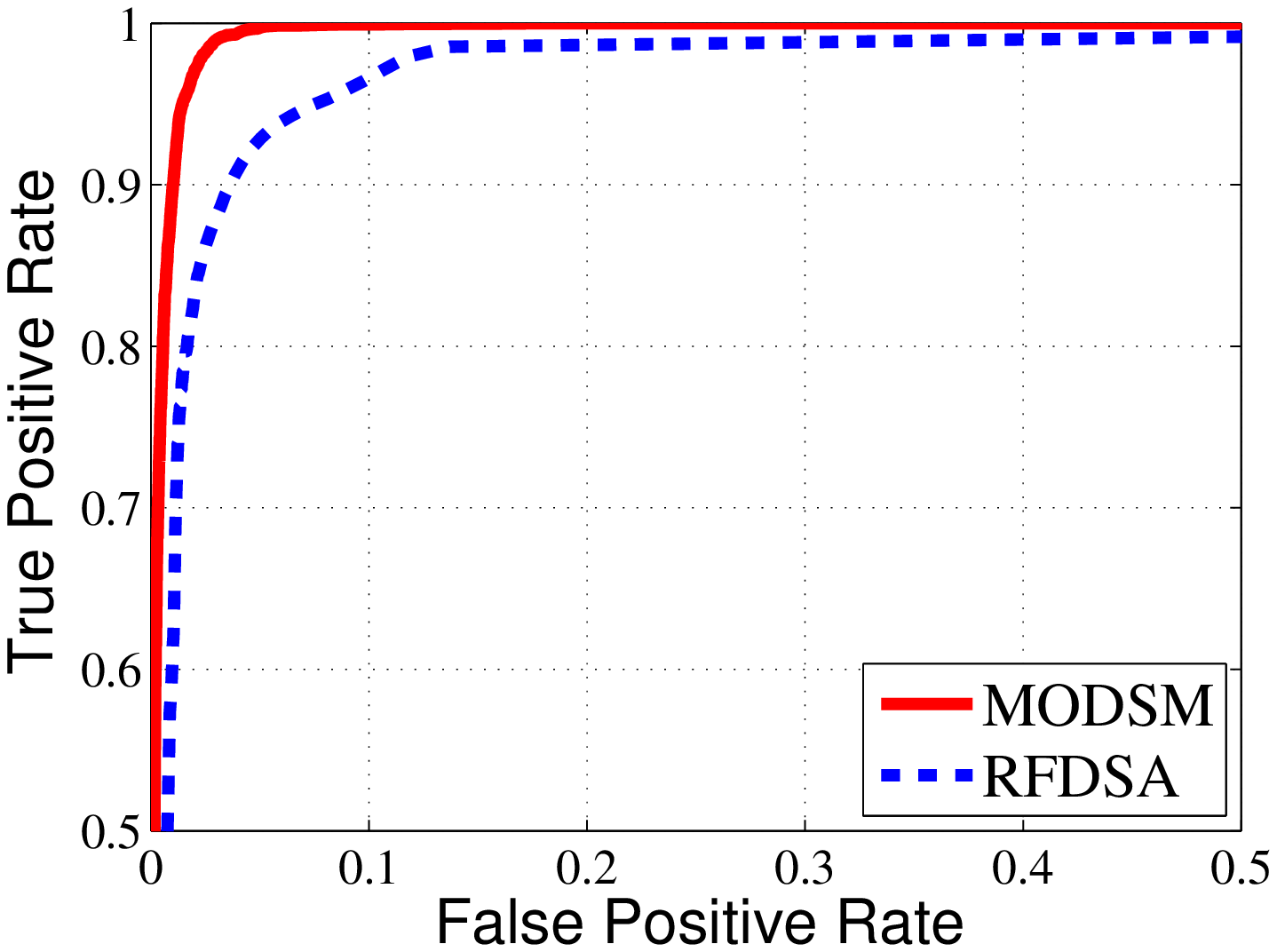}}
\subfloat[]{\includegraphics[width=0.5\linewidth]{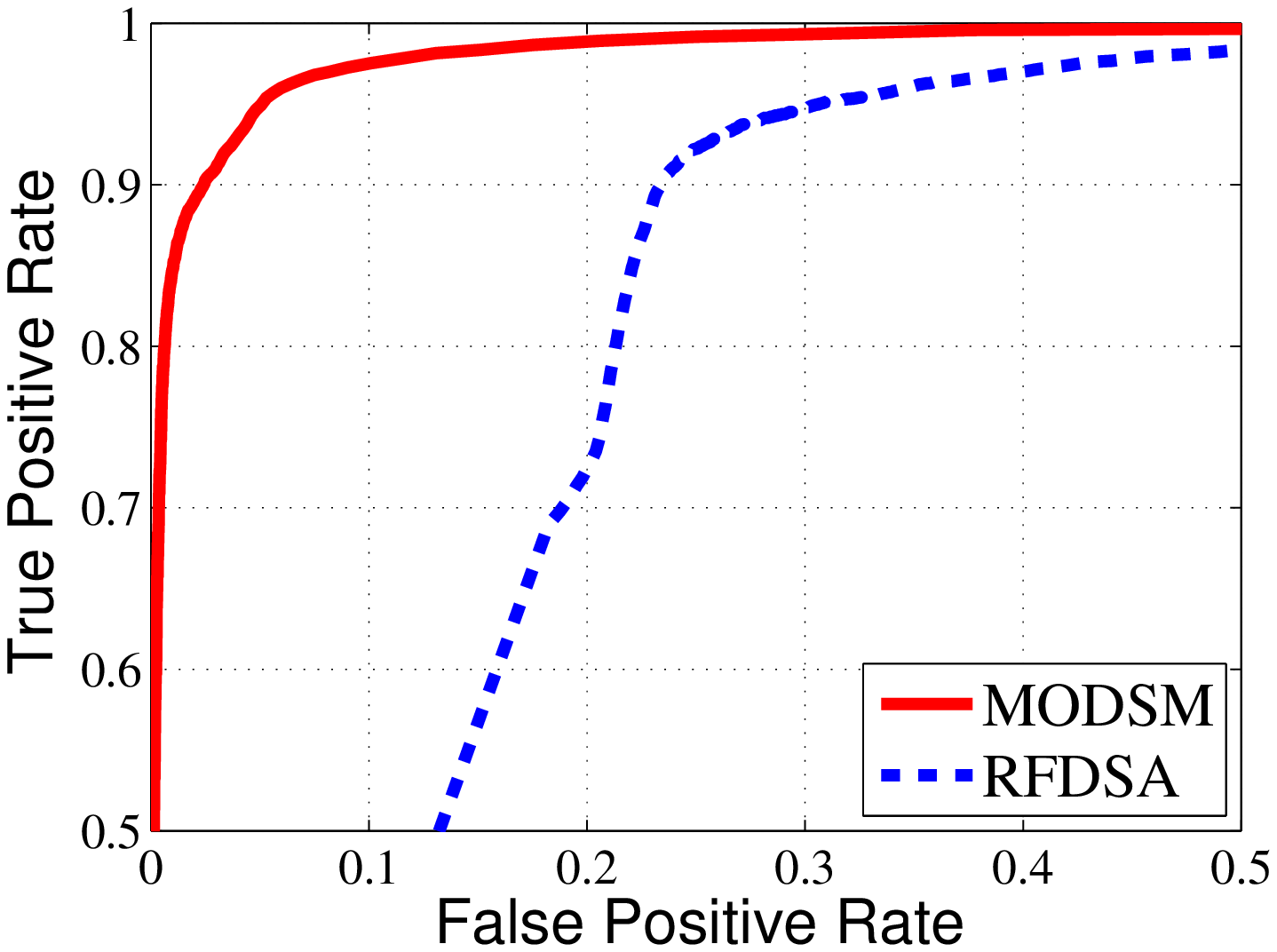}}
\caption{ROC curves on the Water Surface, Escalator, Fountain, and Campus video.}
\label{fig.8}
\end{figure}

The ROC curves of the MODSM and RFDSA on the Water Surface, Escalator, and Fountain, and Campus videos are shown in Fig. {\ref{fig.8}} where the superiority of the MODSM can be observed. Take the Fountain video as an example. The true positive rates (i.e., recall) of MODSM and RFDSA are respectively 0.99 and 0.935 when the false positive rate is 0.05. Note that the DOCOLOR method cannot generate the ROC curves because of their binary values of the estimated foreground and background.

\begin{figure}[!t]
\centering
\subfloat{\includegraphics[width=1\linewidth]{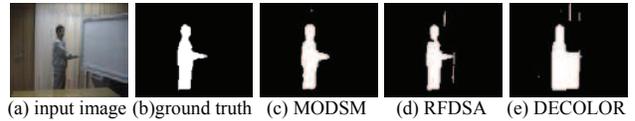}}
\caption{Detected objects for a frame of the Curtain video.}
\label{fig.9}
\end{figure}

\begin{figure}[!t]
\centering
\subfloat{\includegraphics[width=1\linewidth]{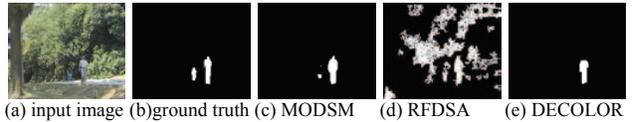}}
\caption{Detected objects for a frame of the Campus video.}
\label{fig.10}
\end{figure}

\begin{figure}[!t]
\centering
\subfloat{\includegraphics[width=1\linewidth]{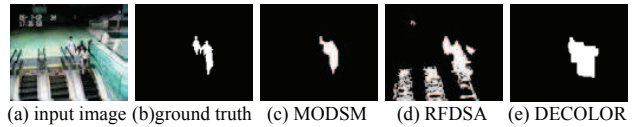}}
\caption{Detected objects for a frame of the Escalator video.}
\label{fig.11}
\end{figure}

Several specific results of MODSM, RFDSA, and DECOLOR are visualized in Fig. {\ref{fig.9}}, Fig. {\ref{fig.10}}, and Fig. {\ref{fig.11}} where (a), (b), (c), (d), and (e) are the current input frame, ground truth of the moving objects, the detected results of MODSM, RFDSA, and DECOLOR, respectively.

Fig. {\ref{fig.9}} (a) is a frame of the Curtain video. Fig. {\ref{fig.9}} (d) shows that RFDSA incorrectly regards the variation caused by motion of the curtain as moving objects and RFDSA results in incomplete neck of the person. Fig. {\ref{fig.9}} (e) shows that DECOLOR gives rise to even more false alarms. Investigating Figs. {\ref{fig.9}} (c) and (b), one can find the result of MODSM is very close to the ground truth.

Fig. {\ref{fig.10}} (a) is a frame of the Campus video. Fig. {\ref{fig.10}} (d) shows that RFDSA incorrectly classifies many waving leafs as meaningful moving objects. Fig. {\ref{fig.10}} (e) tells that DECOLOR cannot detect the left small person and the head of the right large person is also mistakenly classified as background. Fig. {\ref{fig.10}} (c) shows that the proposed method is powerful for classifying the waving leafs as background and detecting both of the persons.

\begin{figure}[!t]
\centering
\subfloat{\includegraphics[width=1\linewidth]{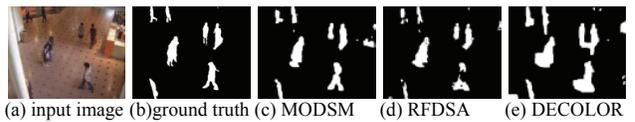}}
\caption{Detected objects for a frame of the Shopping Mall video.}
\label{fig.12}
\end{figure}

Fig. {\ref{fig.11}} (a) is a frame of the Escalator video. Fig. {\ref{fig.11}} (d) shows that RFDSA classifies moving escalator as semantically meaningful moving objects. Because of using the information of saliency map, the proposed MODSM (Fig. {\ref{fig.11}} (c)) avoids the errors of RFDSA. Fig. {\ref{fig.11}} (e) shows that DECOLOR has almost not missed alarms but has many false alarms. The result (Fig. {\ref{fig.11}} (c)) of MODSM is the best among the three methods.

Fig. {\ref{fig.12}} (a) is a frame of the Shopping Mall video. It can be seen that MODSM is comparable and even slightly better than RFDSA and DECOLOR.

\begin{figure}[!t]
\centering
\subfloat{\includegraphics[width=1\linewidth]{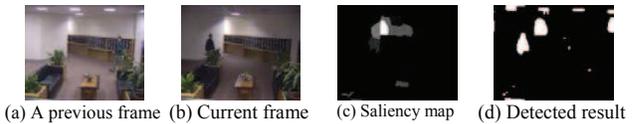}}%
\caption{Detected result for a frame of the Lobby video. }
\label{fig.13}
\end{figure}

As can be seen from Table {\ref{T.3}}, the proposed method MODSM results unsatisfying results on the Lobby video. Fig. {\ref{fig.13}} attempts to explain the reason. On the one hand, switching from light on (Fig. {\ref{fig.13}} (a)) to light off (Fig. {\ref{fig.13}} (b)) gives rise to large variation which is difficult for the basis vectors $\bf U$ to capture. On the other hand, the saliency map is not satisfying on the regions of the moving object (person). In this case, introducing the bad saliency map (Fig. {\ref{fig.13}} (c)) has a negative influence on the task of moving object detection. The research progress of salient object detection is helpful for improving the performance of the propose method.

\section{Conclusion and future work}
In this paper, we have presented a moving object detection method. The method makes use of saliency map by incorporating it into a unified objective function where the properties of sparsity, low-rank, connectivity, and saliency are integrated. The manner of using saliency map yields smaller number of false alarms and missed alarms. Our future work will apply the idea of using saliency map to other state-of-the-art incremental and batch methods of moving object detection. Moreover, we will investigate other state-of-the-art methods of generating saliency map.

\begin{small}

\end{small}

\end{document}